\pdfoutput=1
\documentclass[10pt,twocolumn,letterpaper]{article}
\usepackage{wacv}
\usepackage{times}
\usepackage{epsfig}
\usepackage{graphicx}
\usepackage{amsmath}
\usepackage{amssymb}
\usepackage{float}
\usepackage{authblk}
\makeatletter
\renewcommand\AB@affilsepx{ , \protect\Affilfont}
\makeatother

\wacvfinalcopy 

\def\wacvPaperID{****} 

\ifwacvfinal\fi
\begin{document}

\title{DGGAN: Depth-image Guided Generative Adversarial Networks for Disentangling RGB and Depth Images in 3D Hand Pose Estimation}
\author[1]{Liangjian Chen}
\author[2]{Shih-Yao Lin}
\author[3]{Yusheng Xie\thanks{Work done prior to joining Amazon}}
\author[4]{Yen-Yu Lin}
\author[2]{Wei Fan}
\author[1]{Xiaohui Xie}
\affil[1]{University of California, Irvine}
\affil[2]{Tencent America}
\affil[3]{Amazon}
\affil[4]{National Chiao Tung University}
\affil[ ]{\textit {\{liangjc2,xhx\}@ics.uci.edu}}
\affil[ ]{\textit {\{shihyaolin,davidwfan\}@tencent.com}}
\affil[ ]{\textit {yushx@amazon.com}}
\affil[ ]{\textit {lin@cs.nctu.edu.tw }}


\maketitle
\ifwacvfinal\thispagestyle{empty}\fi

\begin{abstract}


Estimating $3$D hand poses from RGB images is essential to a wide range of potential applications, but is challenging owing to substantial ambiguity in the inference of depth information from RGB images.
State-of-the-art estimators address this problem by regularizing $3$D hand pose estimation models during training to enforce the consistency between the predicted $3$D poses and the ground-truth depth maps.
However, these estimators rely on both RGB images and the paired depth maps during training.
In this study, we propose a conditional generative adversarial network (GAN) model, called Depth-image Guided GAN (DGGAN), to generate realistic depth maps conditioned on the input RGB image, and use the synthesized depth maps to regularize the $3$D hand pose estimation model, therefore eliminating the need for ground-truth depth maps. 
Experimental results on multiple benchmark datasets show that the synthesized depth maps produced by DGGAN are quite effective in regularizing the pose estimation model, yielding new state-of-the-art results in estimation accuracy, notably reducing the mean $3$D end-point errors (EPE) by $4.7\%$, $16.5\%$, and $6.8\%$ on the RHD, STB and MHP datasets, respectively.  

\end{abstract}


\section{Introduction}
\label{sec:intro}

\begin{figure}[t]
\hspace{-0.3cm}
\begin{tabular}{cc}
\centering 
\fbox{
\includegraphics[width=0.2\textwidth]{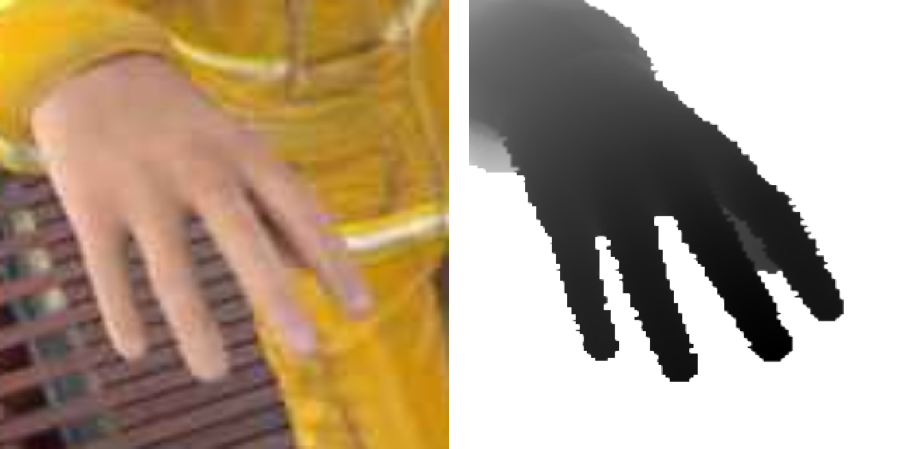}\label{fig:pair}
}
&
\fbox{
\includegraphics[width=0.2\textwidth]{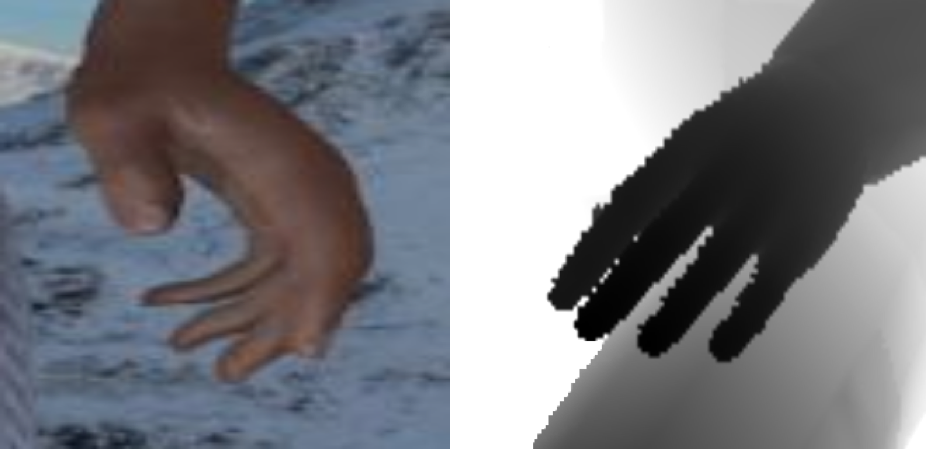}\label{fig:unpair}
}
\\
(a)&(b)
\end{tabular}
\caption{Training examples in a generic $3$D HPE dataset: (a) paired RGB and depth images; (b) unpaired RGB and depth images. Our work does not rely on paired training data and therefore is applicable to both RGB-only and depth-only $3$D HPE tasks.}
\label{fig:rgb_vs_d}
\end{figure}

Vision-based $3$D hand pose estimation ($3$D HPE) aims to estimate the $3$D keypoint coordinates of a given hand image.
$3$D HPE has drawn increasing attention owing to its wide applications to human-computer interaction (HCI)~\cite{antoshchuk2018gesture, lin2013airtouch}, sign language understanding~\cite{zafrulla2011american}, augmented/virtual reality (AR/VR)~\cite{mueller2018ganerated, hung2016re}, and robotics~\cite{antoshchuk2018gesture}.
RGB images and depth maps are two the most commonly used input data for the $3$D HPE task.
An example of a hand image and its corresponding depth map is shown in Figure~\ref{fig:rgb_vs_d}(a).
%
Depth map can provide $3$D information related to the distance of the surface of human hands.
Training networks with depth maps has been proven to achieve significant progress on the $3$D HPE task~\cite{cai2018weakly, iqbal2018hand}. 
In addition, with the depth information provided by the depth maps, the hand segmentation task can be effectively solved.
Unfortunately, capturing depth maps often requires specific sensors (\eg~Microsoft Kinect, RealSense), which limits the usability of those state-of-the-art methods based on depth maps.
Commercial depth sensors are usually much more expensive than RGB cameras.
On the other hand, RGB images are the most commonly used input data in the HPE task because it can be easily captured by abundant low-cost optical sensors such as webcams and smartphones.
However, $3$D HPE from RGB images is a challenging task. 
%


%
In the absence of depth information, estimating $3$D hand pose from a monocular RGB image is intrinsically an ill-posed problem. 
To address this issue, the state-of-the-art methods such as~\cite{cai2018weakly, ge2019handshapepose}
leverage both RGB hand images and their paired depth maps for the $3$D HPE task.
Their $3$D hand pose inference process takes an RGB image and the paired depth information into account.
They first regress $3$D hand poses on RGB images, and then utilize a separate branch to regularize the predicted $3$D hand pose by using the paired depth maps. 
The objective of the depth regularizer is to make the predicted $3$D keypoint positions consistent with the provided depth map. 
%
%
It results in two major advantages: 1) training networks with depth maps can efficiently improve the hand pose estimator by using the depth information to reduce the ambiguity and 2) enabling $3$D HPE based on merely RGB images during the inference stage.
%
These approaches require paired RGB and depth training images.
Unfortunately, most existing hand pose datasets only contain either depth maps or RGB images, instead of both. 
It makes the aforementioned approaches not applicable to such datasets. 
Besides, the unpaired RGB and depth training images cannot be explorited for them. 
Figure~\ref{fig:rgb_vs_d}(b) shows an example of unpaired RGB and depth map images.
 
To tackle this problem, we propose a novel generative adversarial networks, called {\em Depth-image Guided GAN} (DGGAN).
Our network contains two modules: {\em depth-map reconstruction} and {\em hand pose estimation}.
The main idea of our approach is to directly reconstruct the depth map from an input RGB hand image in the absence of paired RGB and depth training images.
Given an RGB image, our depth-map reconstruction module aims to infer its depth map.
Our hand pose estimation module takes RGB and depth information into account to infer the $3$D hand pose.
In the hand pose estimation module, we infer the $2$D hand keypoints on the input RGB image, and regress the $3$D hand pose by using the inferred 2D keypoints.
The depth map is then used to regularize the inferred $3$D hand pose.
Unlike most existing $3$D HPE models, the real depth maps used to train our DGGAN model do not require any paired RGB images.
Once DGGAN is learned, the proposed HPE module directly infers the hand pose by using an RGB image and guided (regularized) by a DGGAN-inferred depth map.
Since the depth-map can be inferred by our depth-map reconstruction module, the proposed DGGAN no longer requires paired RGB and depth images.
%
%
Our DGGAN jointly trains the two modules in an end-to-end trainable network architecture.  
Experimental results on multiple benchmark datasets demonstrate that our DGGAN not only reconstructs the depth map of an input RGB image, but also significantly improves the $3$D hand pose estimator via an additional depth regularizer.

%

 
The main contributions of this study are summarized as follows:
 \begin{enumerate}
    \item We propose a depth-map guided adversarial neural networks (DGGAN) for $3$D hand pose estimation from RGB images. Our network can jointly infer the depth information from input RGB images and estimate the 3D hand poses. 
     \item We introduce a depth-map reconstruction module to infer the depth maps from input RGB images while learning to predict $3$D hand poses. Our DGGAN is trained on readily accessible hand depth maps that are not paired with RGB images.
     \item Experimental results demonstrate that our approach achieves new state-of-the-art in $3$D hand pose prediction accuracy on three benchmark datasets, including the RHD, STB, and MHP datasets.
 \end{enumerate}
\section{Related Work}
\label{sec:related}

Research topics related to this work are discussed below.

\subsection{3D HPE from Depth Images}

$3$D HPE from depth mapshas been extensively studied.
Existing approaches in this field make noticeable advances~\cite{wan2018dense, Yuan_2018_CVPR, deng2017hand3d, Wu18HandPose,ge2018hand, Ge_2018_ECCV,li2018point}. 
Wan~\etal~\cite{wan2018dense} propose a dense regression approach to fit the parameters of a deformed hand model. 
Ge~\etal~\cite{ge2018hand, Ge_2018_ECCV} present PointNet\cite{Qi_2017_CVPR} to extract hand features and regress hand joint locations by referring to the extracted features. 
Wu~\etal~\cite{Wu18HandPose} adopt the intermediate dense guidance map supervision to generate hand heatmaps. 
Although the existing methods achieve very accurate estimation results, they typically rely on the hand data captured by high-precision depth sensors, which are still expensive to have in practice and usually require data collection in a lab environment.
Different from the models in the aforementioned methods,  our model performs inference on RGB data without the need of depth maps.

\begin{figure*}[h]
	\begin{center}
		\begin{tabular}{c}
		\includegraphics[width=0.7\textwidth]{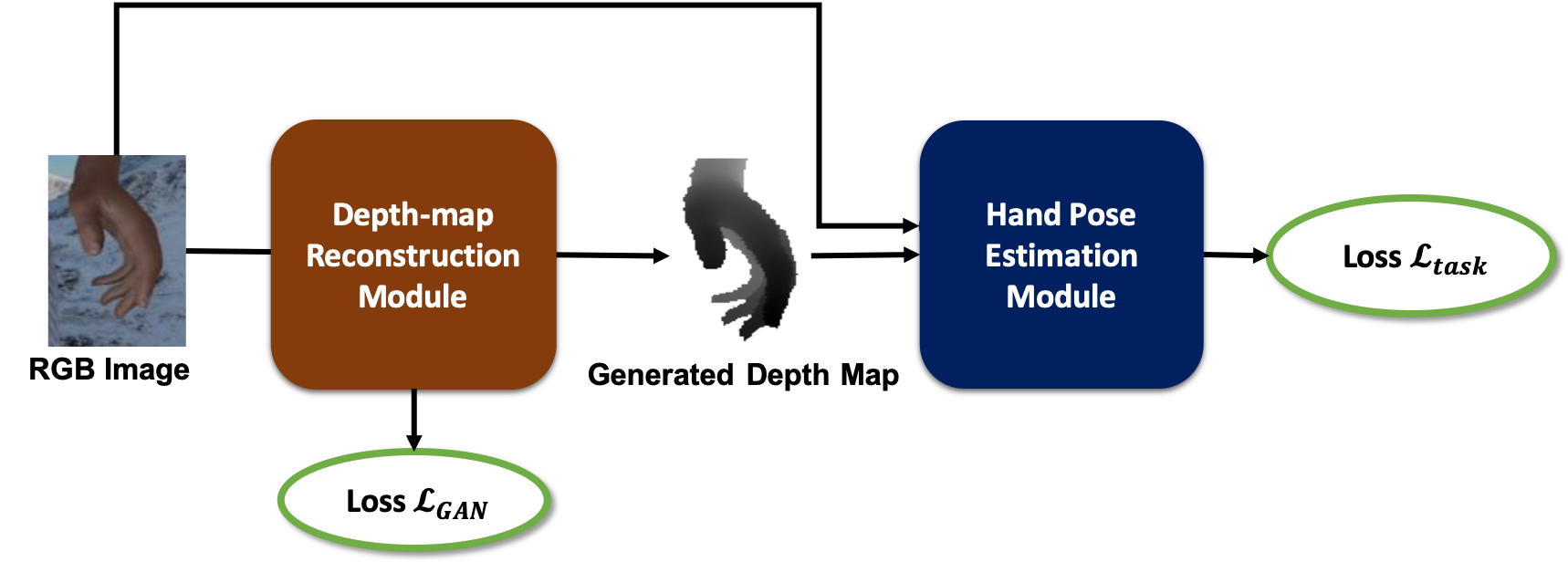}
		\end{tabular}
	\end{center}
	\caption{Overview of the proposed DGGAN. 
	DGGAN consists of two modules, a {\em depth-map reconstruction module} shown in Figure~\ref{fig:task} and a {\em hand pose estimation module} shown in Figure~\ref{fig:task}. 
	The former module trained using the GAN loss aims at inferring the depth map of a hand based on the input RGB image and making the generated depth map looks realistic.
	The latter module trained using the task loss estimates hand poses from the input RGB and the GAN-reconstructed depth images.
	%
	}
	\label{fig:GAN}
\end{figure*}

\subsection{3D HPE from Monocular RGB Images}

Due to the wide availability of RGB cameras, $3$D HPE from monocular RGB images is becoming increasingly popular in computer vision applications. 
Many recent methods aim at estimating hand joint locations directly from a single RGB image \cite{cai2018weakly,iqbal2018hand,ge2019handshapepose,zb2017hand,mueller2018ganerated,chen2018generating,yang2018disentangling,boukhayma20193d,zhe2019Geo, tekin2019h+}. 
Zimmermann \etal \cite{zb2017hand} use $2$D convolutional neural networks (CNN) to extract features from an RGB image, and regress the $3$D hand joint locations. 
However, their method suffers from depth ambiguity due to the absence of depth information. 
Developing the methods upon the work by Zimmermann~\etal, Iqbal~\etal~\cite{iqbal2018hand} and Cai~\etal~\cite{cai2018weakly} inherit and adopt a similar $2$D CNN architecture for extracting image features. 
Iqbal~\etal use depth maps as intermediate guidance while Cai~\etal treat depth maps as a regularizer in a weakly supervised manner. 
Though these two methods make substantial progress in terms of estimation accuracy, there currently exist few datasets that fulfill their requirement of paired depth maps and RGB images.
Ge~\etal~\cite{ge2019handshapepose} take one step further by predicting the hand mesh from an RGB image and then the 3D hand joint locations based on the mesh. 
However, their method requires paired mesh information which is even rarer among all existing datasets. 

Compared with these methods, our method also uses depth information during training, but it does not require any paired RGB images and depth maps. 
Thus, it is much more flexible since it can consume RGB images and depth maps from different datasets or sources.

\subsection{3D Mesh Estimation from RGB Images}

To further enhance $3$D HPE~\cite{baek2019pushing,boukhayma20193d,ge2019handshapepose,joo2018total}, hand mesh estimation can be included. %
Namely, the model estimates not only the hand joints but also the hand surface mesh. 
However these methods such as \cite{ge2019handshapepose} have a common drawback: 
They require additional mesh annotations which are even more expensive to obtain than joint locations. 
Thus, they are typically trained on synthetic datasets due to this limitation. 
Seungryul~\etal~\cite{boukhayma20193d} introduce an iterative learning method to refine mesh shapes and achieve very good performance. 
However, like $3$D hand joint locations, hand meshes highly rely on additional supervision from hand segment maps which are typically not available in nowadays hand pose datasets. 
The method by boukhayma~\etal~\cite{boukhayma20193d} is the only extra-data-free method, but its performance is limited.

\subsection{GAN-based Image Translation}

Generating images using generative adversarial networks (GAN)~\cite{goodfellow2014generative} has gained remarkable progress. 
Many approaches explore how to better manipulate images by applying GAN models \cite{hoffman2017cycada,isola2017image,zhu2017unpaired,Chen_2019_CrDoCo}. 
Isola~\etal~\cite{isola2017image} propose the Pix2Pix network which translates label or edges maps to synthesized photos, reconstructs objects from edge maps, or colorizes images. 
Zhu~\etal~\cite{zhu2017unpaired} introduce the cycle-consistent generated adversarial network (CycleGAN). %
CycleGAN uses the cycle consistency loss to disentangle the input and output pair and therefore does not need paired input. 
Hoffman~\etal~\cite{hoffman2017cycada} propose cycle-consistent adversarial domain adaptation (CyCADA).
Compared to CycleGAN, CyCADA contains a segmentation loss. %
As a result, CyCADA not only translates images from one modality to another but also deals with a specific visual task. 

Applying the generative adversarial model to RGB hand images for hand pose estimation is also gaining popularity. 
Muller~\etal~\cite{mueller2018ganerated} introduce the geometry consistent GAN (GeoConGAN) to generate synthetic image data for training. 
Chen~\etal~\cite{chen2018generating} propose the tonality-alignment generative adversarial networks (TAGAN) for producing more realistic images from synthetic images for hand pose estimator training. 
However, these methods only focus on generating RGB images. 
None of them generates depth maps for assisting hand pose estimator training.

\section{Our Approach}
\label{sec:method}

Our goal is to estimate the $3$D hand pose from a monocular RGB hand image. 
Although the existing state-of-the-art methods~\cite{boukhayma20193d,icppso, Yuan_2018_CVPR} have shown that training networks with RGB and depth images can improve the $3$D hand pose estimators, few $3$D hand pose datasets consist of paired RGB and depth images.
To deal with the lack of paired data issue, we propose a novel adversarial neural network, called depth-map guided generated adversarial networks (DGGAN) illustrated in Figure \ref{fig:GAN}, which can jointly learn to infer the depth map from an RGB image of hand and to estimate $3$D hand pose.
In the following, we give an overview of the proposed DGGAN and describe the two major modules of DGGAN in detail. 

\subsection{Overview of DGGAN}

The proposed DGGAN consists of two major modules, a {\em depth-map reconstruction} module and a {\em hand pose estimation} module.
Its network architecture is shown in Figure~\ref{fig:GAN}.

Given an RGB hand image $\mathbf{I}$, we want to estimate the $K$ $3$D hand joint locations $\mathbf{J}^{xyz}\in \mathbb{R}^{3 \times K}$. 
Each column in the $3 \times K$ matrix is a vector of size $3$ and represents the $(x,y,z)$ coordinates of a joint, i.e., $\mathbf{J}^{xyz} = [J^{xyz}_1, J^{xyz}_2 , \ldots , J^{xyz}_K ]$. 


\begin{figure}[t]
\hspace{-0.5cm}
		\begin{tabular}{c}
		\includegraphics[width=0.48\textwidth]{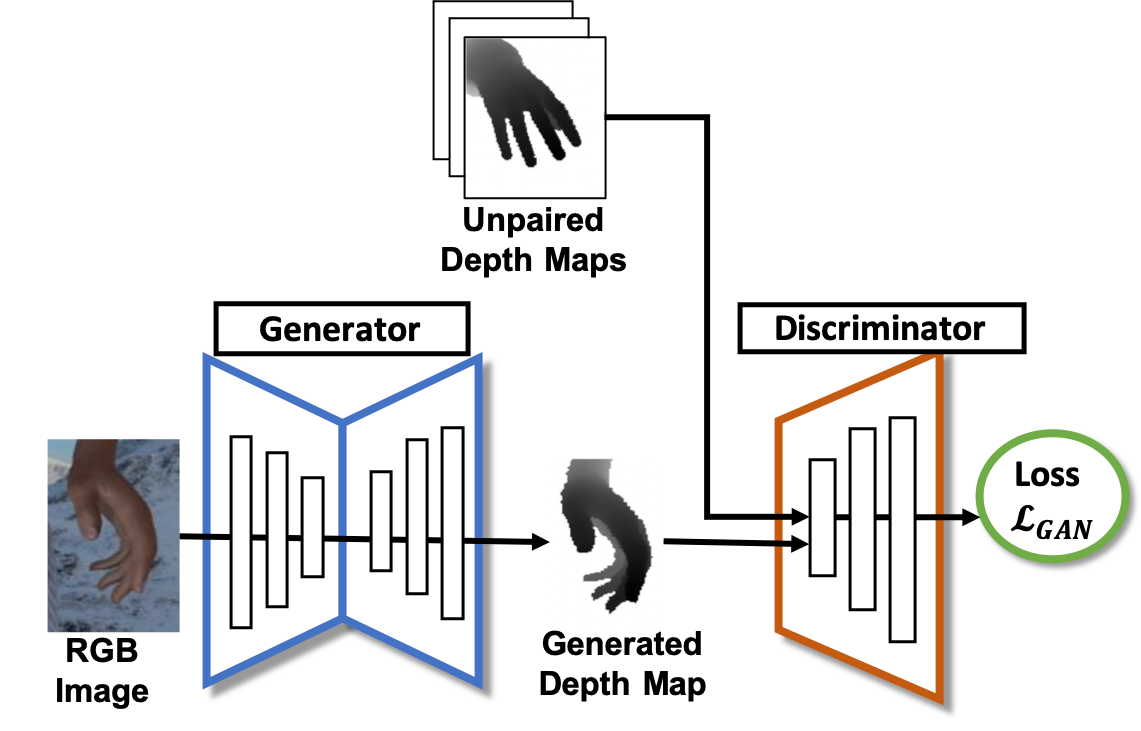}
		\end{tabular}
	\caption{Network architecture of the depth-map reconstruction module.}
	\label{fig:depth_module}
\end{figure}

The two modules in the proposed DGGAN $G$ are trained by using the GAN loss $\mathcal{L}_{GAN}$ and the task loss  $\mathcal{L}_{task}$, respectively.
The objective of learning $G$ is formulated as a min-max game:
\begin{equation}
   G^* = \arg \min_{G}\max_{D}(\lambda_{t} \mathcal{L}_{task} + \lambda_{g} \mathcal{L}_{GAN}), 
\end{equation}
where $\lambda_{t}$ and $\lambda_{g}$ control the relative importance of these two loss terms.


\begin{figure*}[t]
	\begin{center}
		\begin{tabular}{c}
		\includegraphics[width=0.9\textwidth]{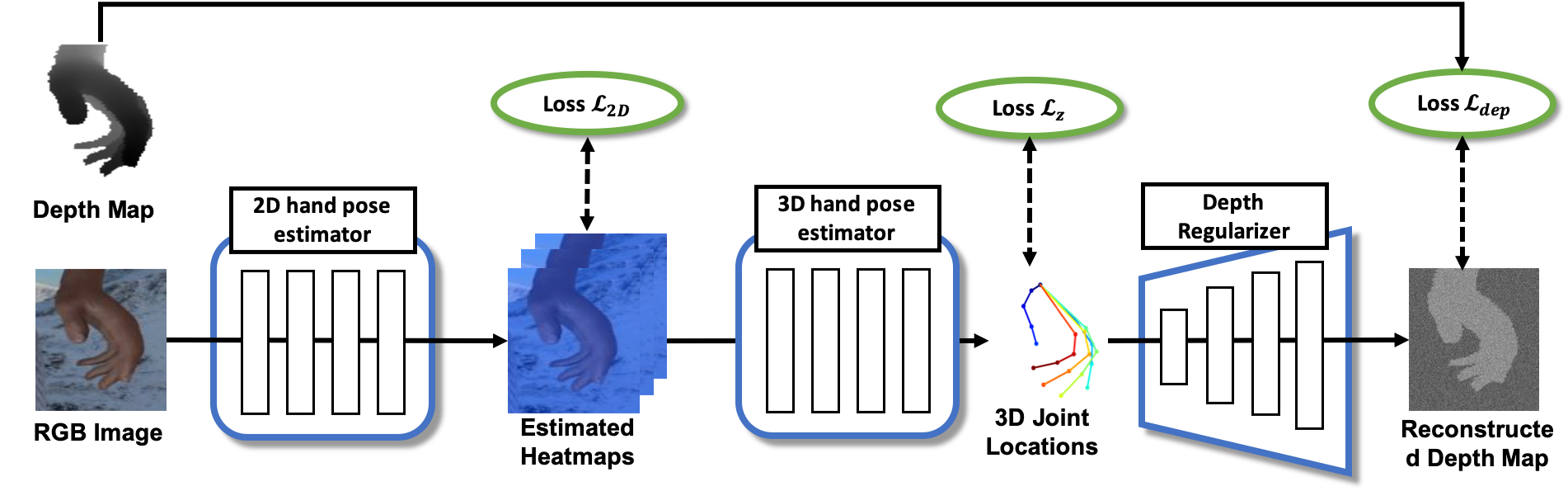}
		\end{tabular}
	\end{center}
	\caption{Architecture of the hand pose estimation module. 
	This module takes paired RGB images and inferred depth maps as inputs. 
	$2$D CPM consumes an RGB image as input and produces the hand joint heatmap. 
	The joint heatmap is fed to the regression network to estimate the $3$D joint locations with the aid of a depth regularizer. 
	The depth regularizer reconstructs the depth map from $3$D joint locations and is trained using L1 loss and the GAN-synthesized depth map as guidance.}
	\label{fig:task}
\end{figure*}

Given an RGB hand image, our depth-map reconstruction module tries to generate its corresponding depth map.
A set of unpaired training depth images is adopted to train the depth-map reconstruction module so that its inferred depth maps are similar to real ones.
To achieve that, the discriminator in this module works on distinguishing real depth maps from fake (generated) ones.
Section~\ref{sec:depth_module} describes the details of depth-map reconstruction.
%
%
%
The depth map inferred from the depth-map reconstruction module together with the input RGB image are fed to the hand pose estimation module for estimating the $3$D hand pose.
In the hand pose estimation module, the input RGB image is used to regress the $3$D hand pose.
The inferred depth-map is adopted to regularize the predicted $3$D hand pose.
The loss for hand pose estimation $\mathcal{L}_{task}$ is adopted for optimization.
Section~\ref{sec:3dpose} describes the details.

\subsection{Depth-map Reconstruction Module}
\label{sec:depth_module}

The depth-map reconstruction module aims at relaxing the requirement of paired RGB and depth images during training.
This module is constructed via an adversarial network that infers the depth map according to an input RGB image.
Figure~\ref{fig:depth_module} shows the network architecture of this module.
In the training phase, our network requires both depth and RGB training images.
Nevertheless, the RGB and depth images do not need to be paired.
We consider the process of inferring depth map from its corresponding RGB image as an unsupervised adaptation problem, where the RGB modality $S$ and depth modality $T$ are both provided. 
We are given a set of RGB images $X_S$ and a set of real depth maps $X_T$.
To translate from $S$ to $T$, we adopt an encoder-decoder architecture $G_{S\rightarrow T}$.
The generator $G_{S\rightarrow T}$ is trained to generate a realistic depth map to fool the discriminator $D$ while $D$ id derived to distinguish the real data $x_t$ and generated fake data $G_{S\rightarrow T}(x_s)$.
%
The loss for the depth-reconstruction modules is as follows:
\begin{equation}
\begin{split}
\mathcal{L}_{GAN}&(G_{S\rightarrow T}, D, X_S, X_T) = \\ &\mathbb{E}_{x_t\sim X_T}[\log D(x_t)] + \\ 
&\mathbb{E}_{x_s\sim X_S}[\log(1 -  D(G_{S\rightarrow T}(x_s)))].
\end{split}
\end{equation}

This loss also provides semantic constraints to force the generator to produce more realistic depth maps. 
By taking as input unpaired RGB and depth images, our depth-map reconstruction module becomes applicable to vastly more hand pose datasets.
%
Furthermore, we can train the network with a large amount of unpaired RGB and depth images.

\subsection{Hand Pose Estimation Module}
\label{sec:3dpose}

Given an inferred depth map computed by the depth-map reconstruction module, we combine it with the input RGB image and feed both to the hand pose estimation module.
The network architecture of the hand pose estimation module is shown in Figure~\ref{fig:task}.
The hand pose estimation module calculates the task loss $\mathcal{L}_{task}$, which is composed of two terms $\mathcal{L}_{task} = \mathcal{L}_{2D}+\mathcal{L}_z$.
The $3$D hand regression loss $\mathcal{L}_{2D}$ and depth regularization loss $\mathcal{L}_z$ are described in section~\ref{sec:regrssion} and~\ref{sec:depth}, respectively.

\subsubsection{3D Hand Pose Regression}
\label{sec:regrssion}



Previous studies~\cite{cai2018weakly} show that depth information can be used to build a powerful regularizer. 
We leverage the depth regularizer for improving the result of $3$D HPE. 
Unlike most previous works where the ground-truth depth maps are needed, our model uses a synthetic depth map generated by the depth-map reconstruction module. 
Our experimental results show that training with such synthetic depth maps substantially helps improve the result of direct regression. 

$3$D hand pose regression takes an RGB image and an inferred depth map as input and outputs joint locations in two steps. 
%
In the first step, we adopt a popular variant of the CPM architecture~\cite{cao2017realtime, wei2016convolutional} as the $2$D joint location predictor. 
This predictor consists of six stages. 
Each stage contains seven convolutional layers followed by a Rectified Linear Unit (ReLu). 
It predicts $K$ heatmaps $\{H_s^k\}_{k=1}^K$ for $K$ different hand joints. 
The pixel value in $k^{th}$ heatmap at stage $s$, $H^k_s$, indicates the confidence that the $k^{th}$ joint is located at this position. 
Following the convention \cite{wei2016convolutional}, the ground-truth heatmap is denoted as $\{H_*^k\}_{k=1}^K$.
Each $H^k_*$ is the Gaussian blur of the Dirac-$\delta$ distribution centered at the ground-truth location of $k^{th}$ joint. 
We train this part of Hand Pose module by standard backpropogation and the mean square error (MSE) loss. 
In addition to the MSE loss, we add the intermediate supervision for each stage. 
The final loss for $2$D location prediction is
\begin{equation}
    \mathcal{L}_{2D} =\frac{1}{6K} \sum_{s = 1}^{6} \sum_{k = 1}^K||H_s^k - H_*^k||_F^2.
\end{equation}

In the second step, the regression network takes the heatmap from CPM as input, and outputs the relative depth.
Its architecture is a mini-CPM (one stage instead of six) followed by three fully connected layers. 
${Z} \in \mathbb{R}^{K \times 1}$ denotes the relative depth of each hand joint. 
We employ smooth L1 loss between ${Z}$ and the ground-truth ${Z}^*$. 
The loss of depth regression $\mathcal{L}_z$ is summarized as follows:
\begin{equation}
    \mathcal{L}_{z} =  \frac{1}{K} \sum_{k = 1}^K
    \left\{
    \begin{aligned}
       \frac{1}{2}({Z}_k - {Z}_k^*)^2, \mbox{ if } |{Z}_k - {Z}_k^*| \le 0.5\\
       |{Z}_k - {Z}_k^*|, \mbox{ otherwise}. 
    \end{aligned}
    \right.
\end{equation}


\subsubsection{Depth Regularizer}
\label{sec:depth}

To provide supervision on every pixel on a depth map, we employ the depth regularizer (DR) proposed in \cite{cai2018weakly}. 
The depth regularizer takes the relative depth as input and predicts a relative depth map $D$.
It reshapes $Z \in \mathbb{R}^{K \times 1}$ to a $K \times 1 \times 1$ tensor, which is considered as a $K$-channel image input. 
We then up-sample this image from $K$-channel with resolution $1 \times 1$ to $1$-channel with the original depth map resolution ($n \times m$) through the 6 layers of transposed CNN. 

We take L1 norm between $D$ and the ground-truth relative depth map $D^*$ as depth regularizer loss $\mathcal{L}_{dep}$, i.e.,
\begin{equation}
    \mathcal{L}_{dep} = ||D - D^*||,
\end{equation} 
where $D^*$ is obtained by input depth map $\hat{D}^*$ as follows
\begin{equation}
D^* = \frac{\hat{D}^* - \min \hat{D}^*}{\max \hat{D}^* - \min \hat{D}^*}.
\end{equation} 
Note that, we only use the ground-truth depth map $\hat{D}^*$ during the initialization stage. 
It would be replaced by DGGAN-generated depth maps once the initialization stage ends.

Combining the loss terms described in Section \ref{sec:3dpose} and Section \ref{sec:depth}, we summarize the loss function for the hand pose estimation module as
\begin{equation}
    \mathcal{L}_{task} =\lambda_{z} * \mathcal{L}_{z} + \lambda_{2D} * \mathcal{L}_{2D} + \lambda_{dep} * \mathcal{L}_{dep},
\end{equation}
where $\lambda_{z}$, $\lambda_{2D}$, $\lambda_{dep}$ control the importance of three different loss terms, respectively.

\begin{figure}[t]
	\begin{center}
		\begin{tabular}{ccccc}
		\includegraphics[width=0.085\textwidth]{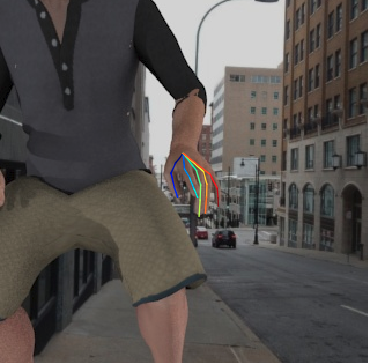}
		\includegraphics[width=0.085\textwidth]{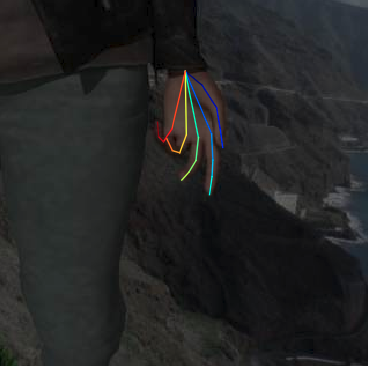}
		\includegraphics[width=0.085\textwidth]{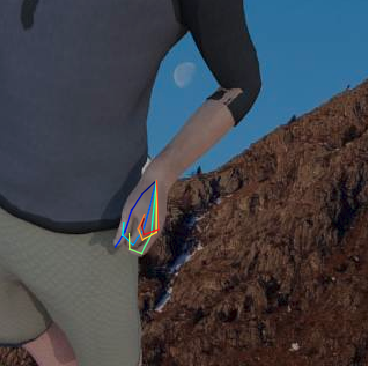}
		\includegraphics[width=0.085\textwidth]{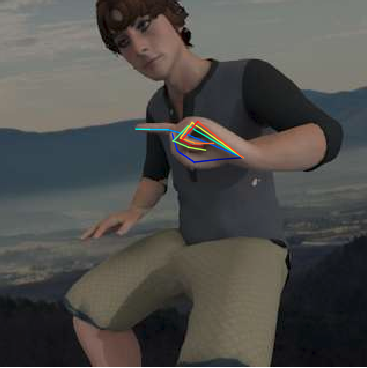}
		\includegraphics[width=0.085\textwidth]{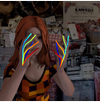}
		\end{tabular}
	\label{fig:rhd}
	\\
	\begin{tabular}{ccccc}
		\includegraphics[width=0.085\textwidth]{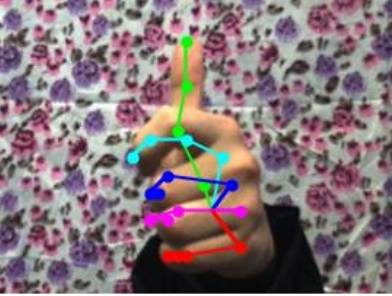}
		\includegraphics[width=0.085\textwidth]{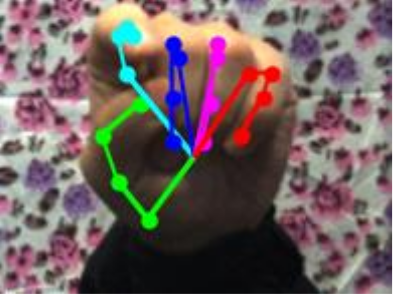}
		\includegraphics[width=0.085\textwidth]{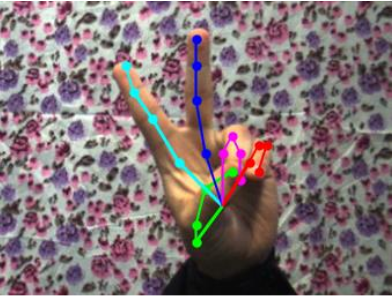}
		\includegraphics[width=0.085\textwidth]{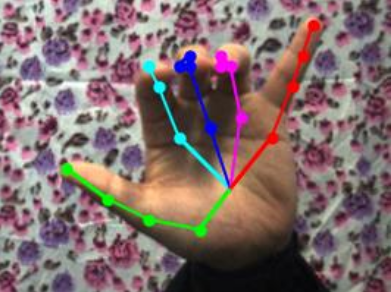}
		\includegraphics[width=0.085\textwidth]{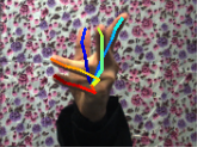}
		\end{tabular}
	\label{fig:stb}
	\\
	\begin{tabular}{ccccc}
		\includegraphics[width=0.085\textwidth]{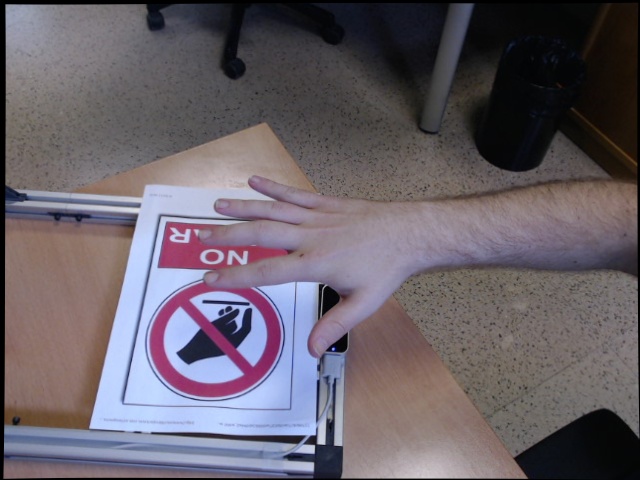}
		\includegraphics[width=0.085\textwidth]{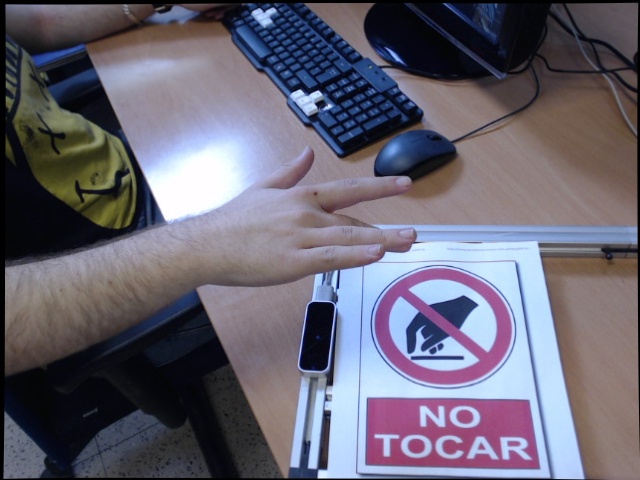}
		\includegraphics[width=0.085\textwidth]{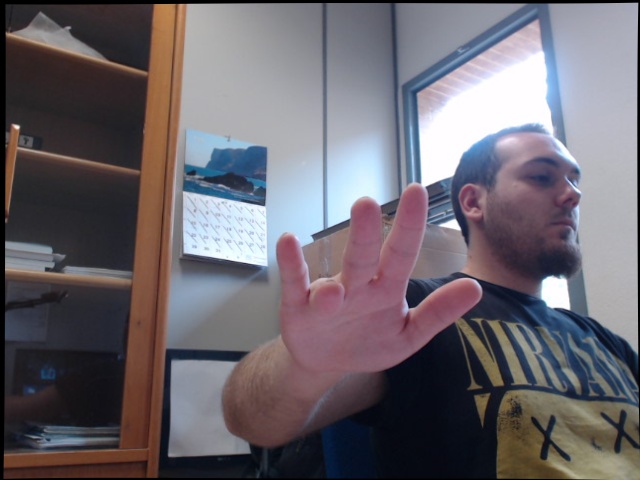}
		\includegraphics[width=0.085\textwidth]{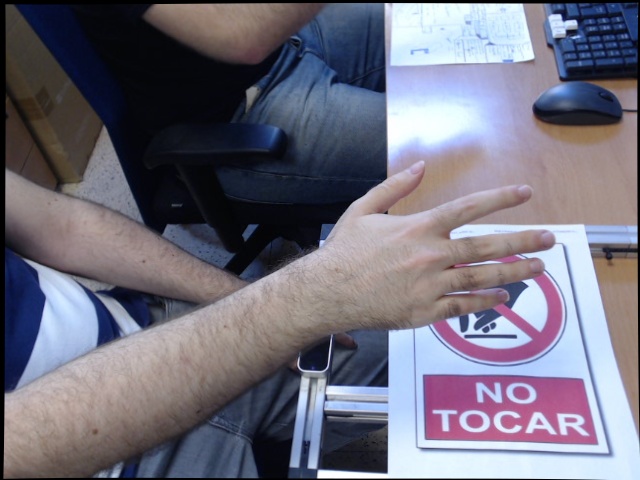}
		\includegraphics[width=0.085\textwidth]{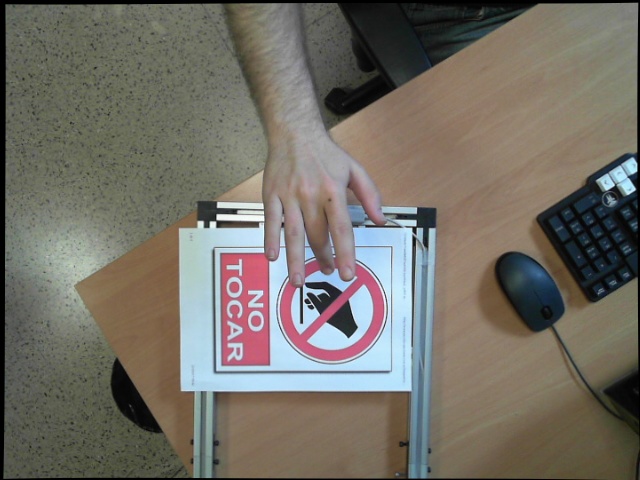}
		\end{tabular}
	\label{fig:cmu}
	\end{center}
	\caption{Some examples of the three benchmark datasets used for evaluation. 
	\textbf{Top Row:} The RHD dataset~\cite{zb2017hand} provides synthetic hand images with $3$D hand keypoint annotations. 
	\textbf{Middle Row:}
	The STB dataset \cite{zhang20163d} contains real hand images with $3$D keypoints. 
	\textbf{Bottom Row:} The MHP \cite{gomez2017large} offers real hand images with $3$D keypoints.}
	\label{fig:dataset}
\end{figure}

\begin{figure*}[]
\begin{tabular}{ccc}
\centering 
\hspace{-0.5cm}
{\includegraphics[width=0.32\textwidth]{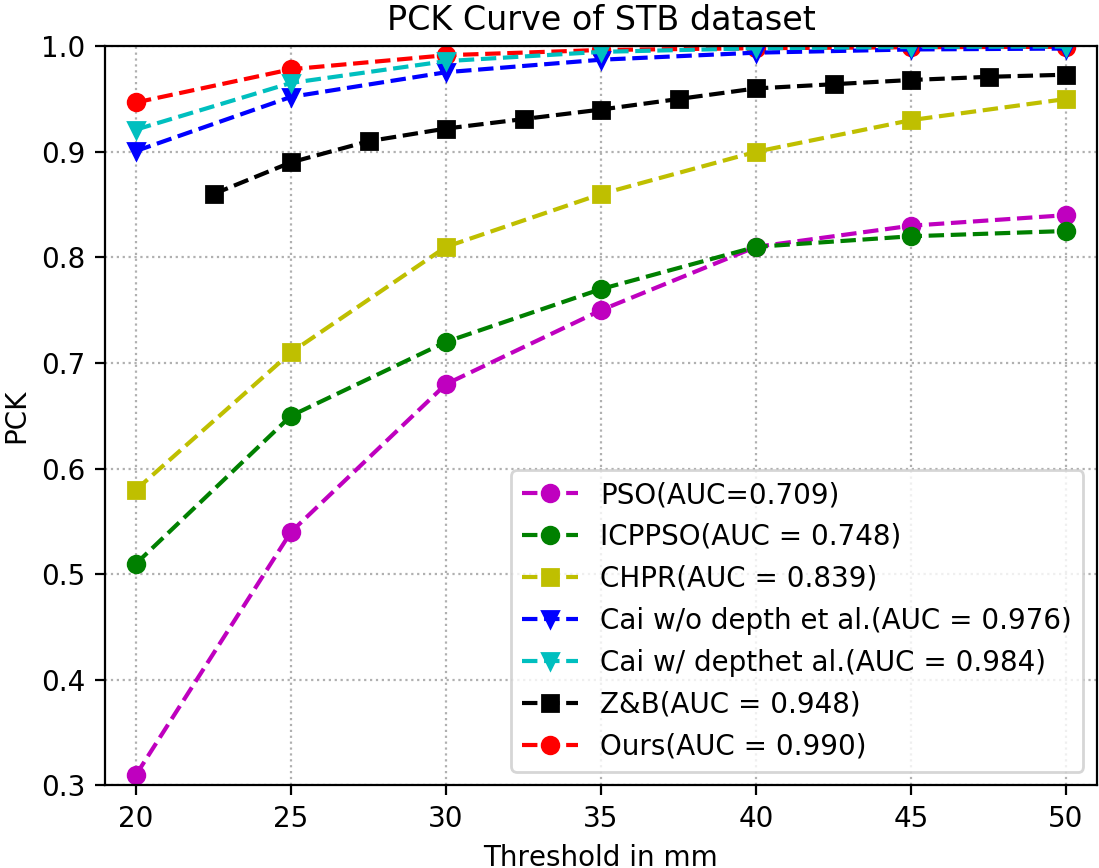}}\label{fig:3DSTB}&
{\includegraphics[width=0.32\textwidth]{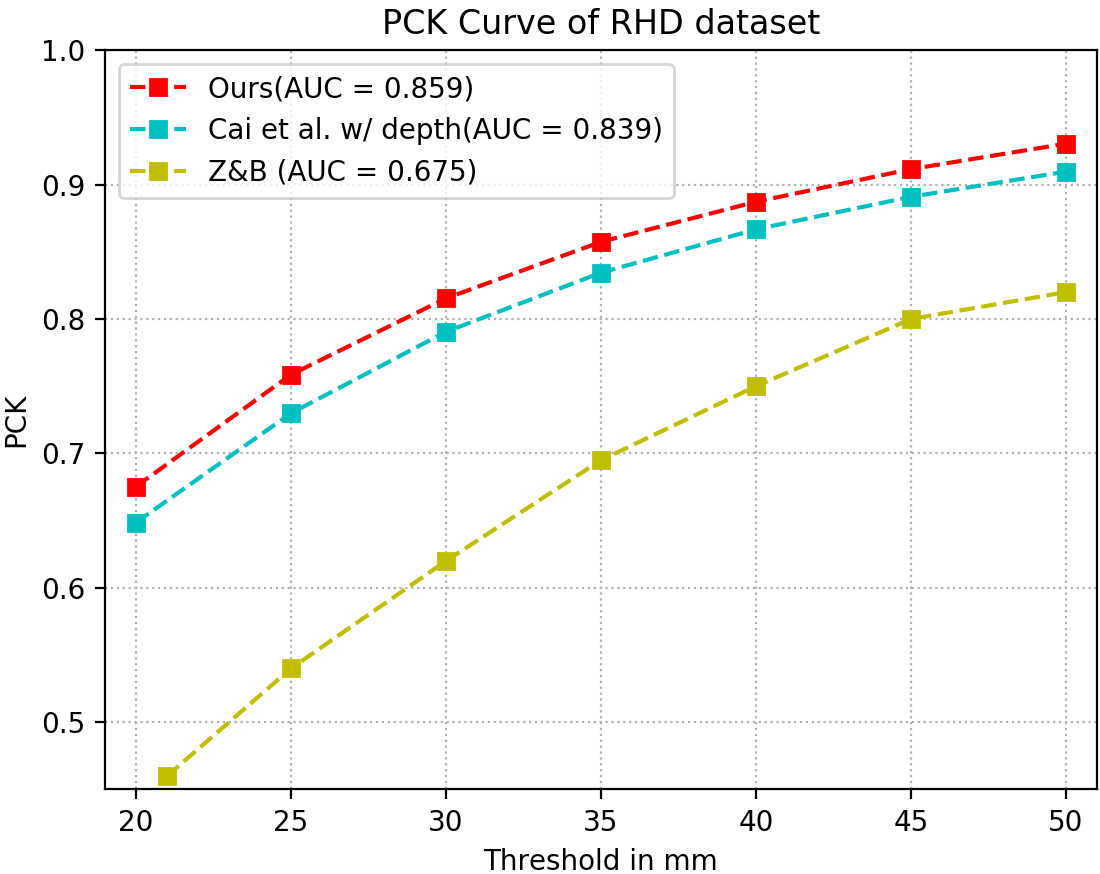}\label{fig:3DRHD}}&
{\includegraphics[width=0.32\textwidth]{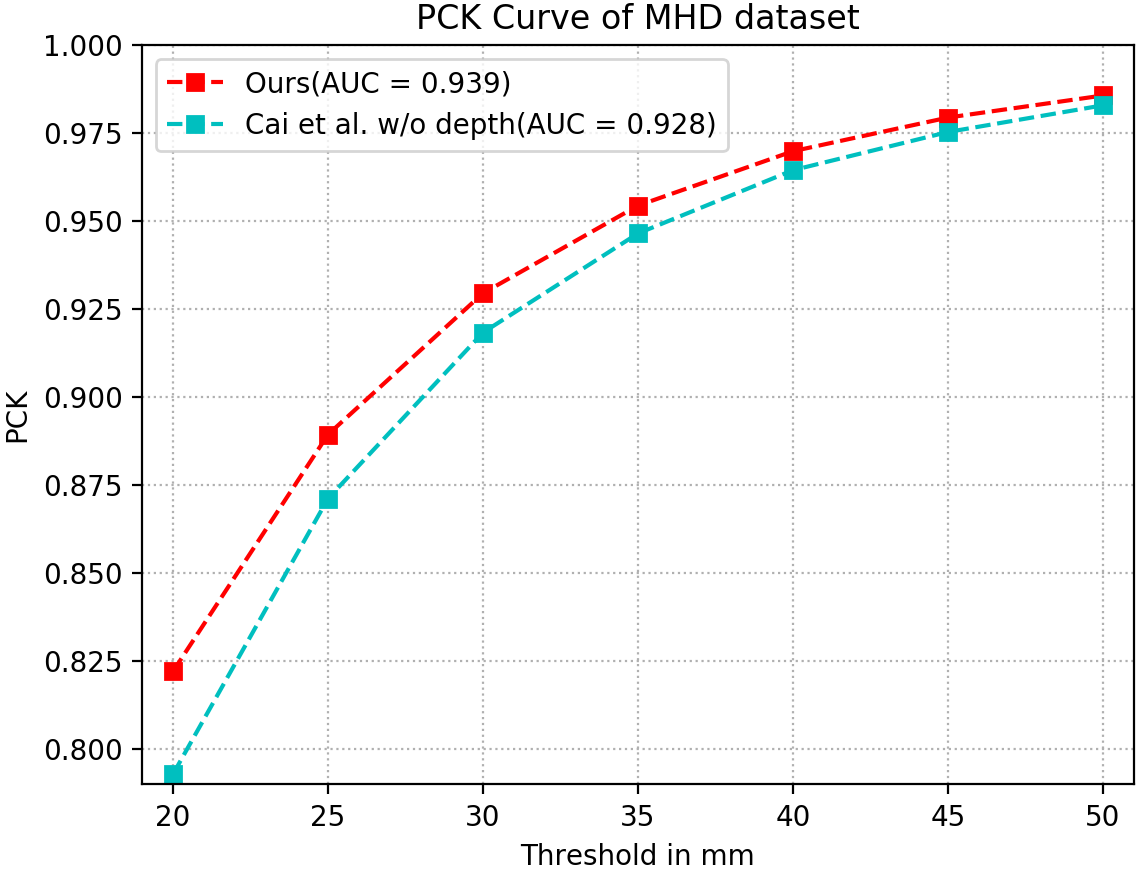}\label{fig:3DMHP}}\\
(a)&(b)&(c)\\
\end{tabular}
\caption{Comparisons with the state-of-the-art approaches on the (a) STB, (b) RHD, and (c) MHP datasets for $3$D hand pose estimation.}
\label{fig:teaser}
\end{figure*}

\section{Experimental Settings}
\label{sec:exp}

This section introduces our experimental settings. 
The selected benchmark datasets for performance evaluation are first given.
The evaluation metric and training details are then presented. 

\subsection{Datasets for Evaluation}
\label{sec:dataset}

We conduct the experiments on three benchmark datasets, including the stereo hand tracking benchmark (STB)~\cite{zhang20163d}, the render hand pose dataset (RHD)~\cite{zb2017hand}, and the multi-view hand pose (MHP) dataset\cite{gomez2017large}.

The STB dataset is a dataset of real hands. 
It contains two different subsets called SK and BB. 
The images in SK are captured by Point Grey Bumblebee2 stereo camera while images in BB are from a depth sensor. 
In our experiments, we use the BB subset for DGGAN training, and leverage the SK subset for unpaired testing.

RHD is a synthetic dataset. 
Zhang~\etal~\cite{zhang20163d} use a $3$D simulator, Maya, to render the images from $20$ different characters doing $39$ actions. 
Each data entry consists of an RGB image and the corresponding depth image, and both $2$D/$3$D annotations.
This dataset is challenging since its images are captured with various view points and of many different hand shapes.

The MHP dataset provides color hand images as well as the bounding boxes of hands and the $2$D and $3$D location of each joint. 
It consists of hand imaegs of $21$ people with different hand movements. 
For each frame, it provides the images from four different angles of view. 
The $2$D and $3$D annotations are obtained by Leap Motion Controller.

Before training, we first crop the hand regions from the original canvas to make sure that hand parts have dominating proportion in the frame. 
Notice that the STB and MHP datasets use the center of a palm rather than a wrist as one of its hand keypoints.
Hence, we revise the annotation to move the center of the palm to the wrist in the same way performed in \cite{cai2018weakly}.

\subsection{Evaluation Metric}

Following the previous works \cite{cai2018weakly,chen2018generating,zb2017hand}, we evaluate the results of hand pose estimation by using 1) the \textit{area under the curve} (AUC) on \textit{percentage of correct keypoints} (PCK) between threshold $20$mm and $50$mm (AUC $20$\_$50$) and 2) the \textit{end-point-error} (EPE): the distance between predicted $3$D joint locations and the ground truth. 
In Table \ref{table:3Dresult}, we report the AUC $20$\_$50$ as well as the mean and the median of EPE over all hand keypoints.

\subsection{Training}

During training, we first initialize the weights of the depth-map reconstruction and hand pose estimation modules in the proposed DGGAN.
Both modules are initialized by fitting the STB dataset (see \ref{sec:dataset}) but trained separately. 
Then, we connect the two modules and fine-tune the whole network in an end-to-end manner. 
For training with the RHD and STB dataset, the discriminator is derived to distinguish the $G_{S\rightarrow T}(x_s)$ and $x_t$, a randomly chosen depth-map from the respective dataset. 
For the MHP dataset, we simply randomly assign a depth-map from RHD dataset as $x_t$ because the MHP dataset does not contain any dense depth maps.

\begin{figure*}[t]
	\begin{center}
		\begin{tabular}{cccccc}
		\includegraphics[width=0.14\textwidth]{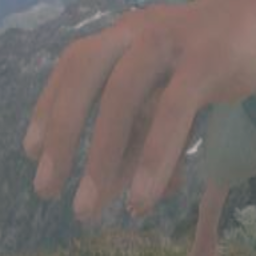}
		\includegraphics[width=0.14\textwidth]{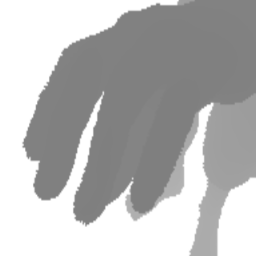}
		\includegraphics[width=0.14\textwidth]{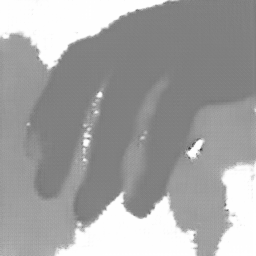}
		\includegraphics[width=0.14\textwidth]{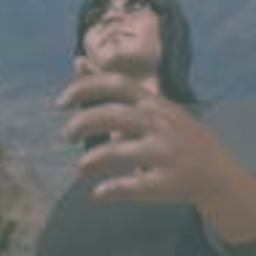}
		\includegraphics[width=0.14\textwidth]{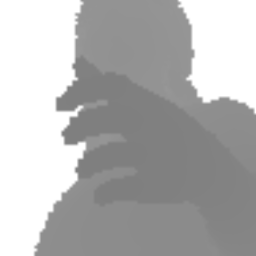}
		\includegraphics[width=0.14\textwidth]{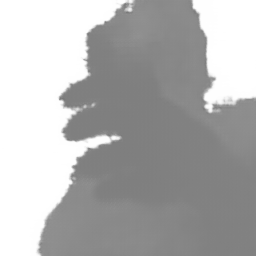}
		\end{tabular}
	\\
	\begin{tabular}{cccccc}
		\includegraphics[width=0.14\textwidth]{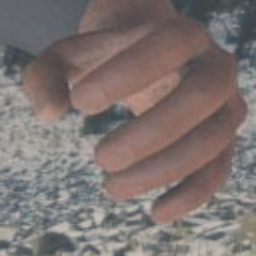}
		\includegraphics[width=0.14\textwidth]{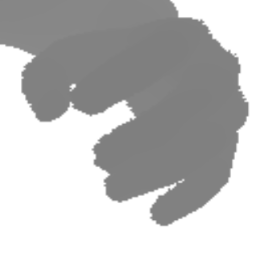}
		\includegraphics[width=0.14\textwidth]{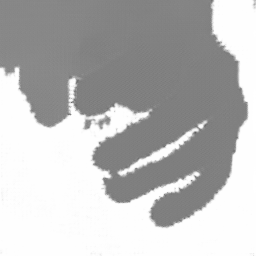}
		\includegraphics[width=0.14\textwidth]{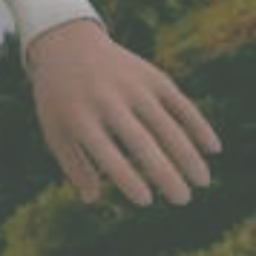}
		\includegraphics[width=0.14\textwidth]{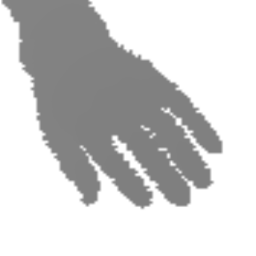}
		\includegraphics[width=0.14\textwidth]{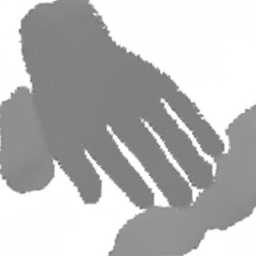}
		\end{tabular}
    \\
    \begin{tabular}{cccccc}
		\includegraphics[width=0.14\textwidth]{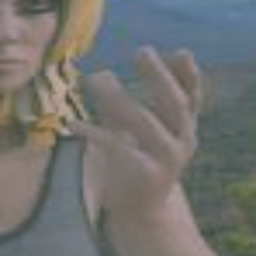}
		\includegraphics[width=0.14\textwidth]{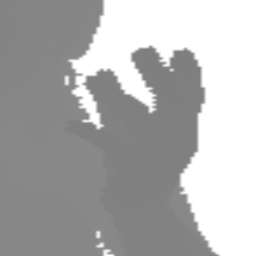}
		\includegraphics[width=0.14\textwidth]{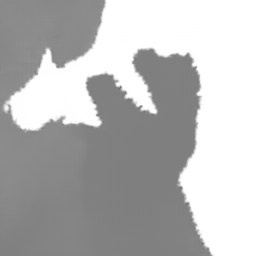}
		\includegraphics[width=0.14\textwidth]{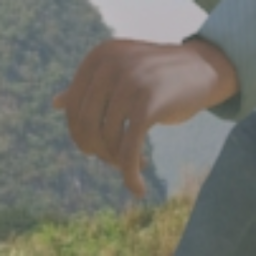}
		\includegraphics[width=0.14\textwidth]{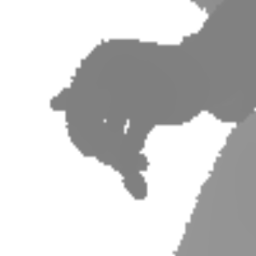}
		\includegraphics[width=0.14\textwidth]{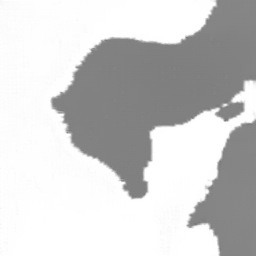}
		\end{tabular}
	\end{center}
	\caption{Comparison between the generated and ground-truth depth maps on the RHD dataset. 
	The first and fourth columns show the RGB images. 
	The second and fifth columns display the real depth maps. 
	The third and sixth columns give the generated depth maps.}
	\label{fig:RHDdepthcmp}
\end{figure*}
\begin{figure*}[]
	\begin{center}
	\begin{tabular}{cccccc}
		\includegraphics[width=0.14\textwidth]{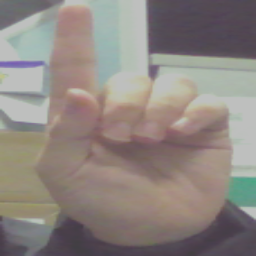}
		\includegraphics[width=0.14\textwidth]{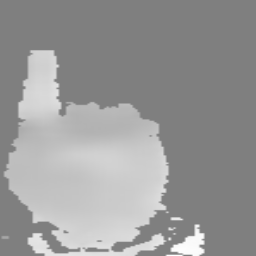}
		\includegraphics[width=0.14\textwidth]{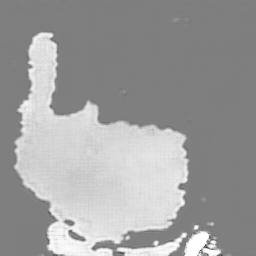}
		\includegraphics[width=0.14\textwidth]{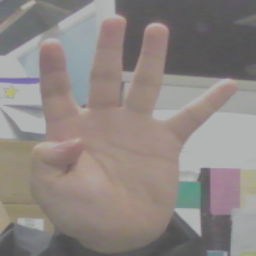}
		\includegraphics[width=0.14\textwidth]{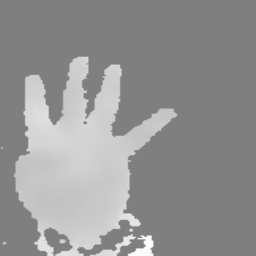}
		\includegraphics[width=0.14\textwidth]{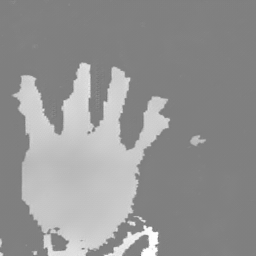}
		\end{tabular}
		\begin{tabular}{cccccc}
		\includegraphics[width=0.14\textwidth]{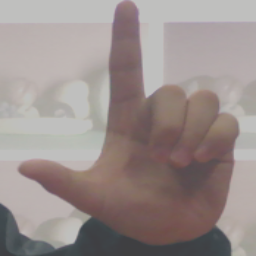}
		\includegraphics[width=0.14\textwidth]{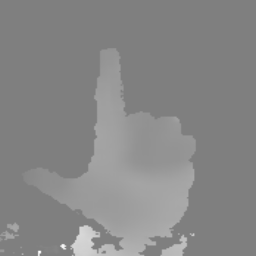}
		\includegraphics[width=0.14\textwidth]{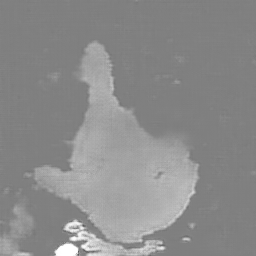}
		\includegraphics[width=0.14\textwidth]{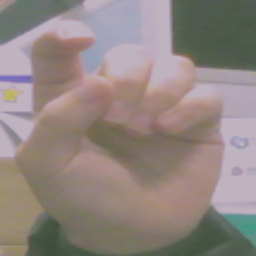}
		\includegraphics[width=0.14\textwidth]{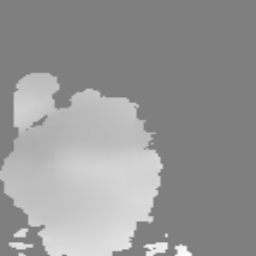}
		\includegraphics[width=0.14\textwidth]{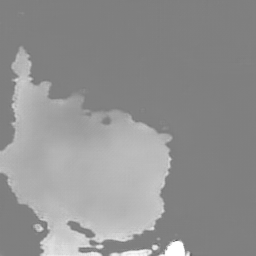}
		\end{tabular}
		\begin{tabular}{cccccc}
		\includegraphics[width=0.14\textwidth]{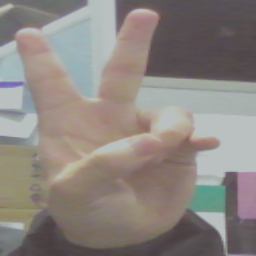}
		\includegraphics[width=0.14\textwidth]{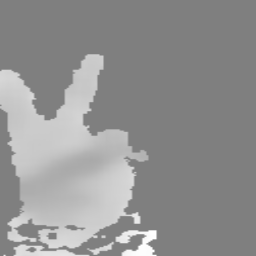}
		\includegraphics[width=0.14\textwidth]{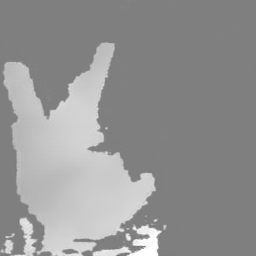}
		\includegraphics[width=0.14\textwidth]{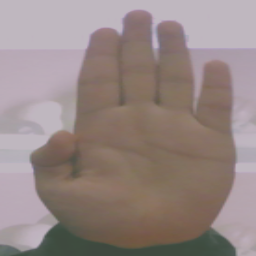}
		\includegraphics[width=0.14\textwidth]{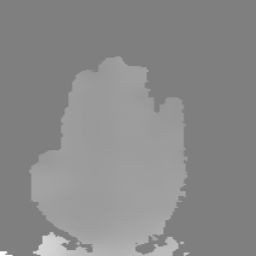}
		\includegraphics[width=0.14\textwidth]{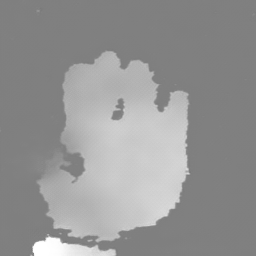}
		\end{tabular}
	\end{center}
	\caption{Comparison between the generated and ground-truth depth maps on the STB dataset. 
	The first and fourth columns show the RGB images. 
	The second and fifth columns display the real depth maps. 
	The third and sixth columns give the generated depth maps.}
	\label{fig:STBdepthcmp}
\end{figure*}
\begin{table*}[t]
\centering
\caption{$3$D pose estimation results on the RHD, STB, MHP datasets. $\uparrow$: higher is better. $\downarrow$: lower is better. \emph{Regression} is the previous State-of-the-art without using paired depth maps. }
    \begin{tabular}{lccc}
    \hline
    & {AUC 20-50} $\uparrow$& {\scriptsize EPE mean (mm)} $\downarrow$& {\scriptsize EPE median (mm)} $\downarrow$\\ 
\hline
    RHD Dataset&&& \\
\hline
    {Regression}                             &0.816       &21.5          & 13.96\\
    {Regression + DR + DGGAN}            &\bf{0.839}  &\bf{19.0}     &\bf{13.17}\\
    {Regression + DR + true depth map}       &0.859       &18.0          & 13.16\\
\hline
    STB Dataset&&& \\
\hline
    {Regression}                            &0.976      & 10.91         & 9.11\\
    {Regression + DR + DGGAN}           &\bf{0.990} & \bf{9.11}     & \bf{7.70}\\
    {Regression + DR + true depth map}       &0.984      & 10.05         & 8.44\\
\hline
    MHP Dataset&&& \\
\hline
    {Regression}                        &0.928      & 14.08         & 10.75\\
    {Regression + DGGAN}           &\bf{0.939} & \bf{13.12}    & \bf{9.91}\\
    \hline
    \end{tabular}

    \label{table:3Dresult}
\end{table*}

\begin{table*}[]
\centering
\caption{EPE mean comparison on the STB dataset between our approach and the method by Boukhayma~\etal~\cite{boukhayma20193d}}
\begin{tabular}{lc}
    \hline
    Method &  { EPE mean (mm)} $\downarrow$ \\ 
\hline
    {Regression + DR + DGGAN (Ours)}           & \bf{9.11}   \\
    {Boukhayma \etal \cite{boukhayma20193d}} &9.76       \\
\hline
\label{table:stoa}
\end{tabular}
\end{table*}

\section{Experimental Results}
\label{sec:res}

For evaluation on the STB dataset, we choose PSO~\cite{pso}, ICPPSO~\cite{icppso}, and CHPR~\cite{chpr} as the baselines. 
In addition, we select the state-of-the-art approaches, Z\&B~\cite{zb2017hand} and that by Cai~\etal~\cite{cai2018weakly} for comparison.

On the RHD dataset, we compare our method with Z\&B~\cite{zb2017hand} and that in~\cite{cai2018weakly}.
Also, on the MHP dataset, we compare our method to that in~\cite{cai2018weakly}.
Note that Cai~\etal~\cite{cai2018weakly} have not released their code yet. 
We re-implement their method and report the results according to our implementation.

\subsection{Ablation Study}

For analyzing the effectiveness of the proposed DGGAN, we conduct ablation studies for DGGAN on three different datasets. 
The detailed results are summarized in Table~\ref{table:3Dresult}. 
Specifically, we conduct the experiments for the following three different settings:
\begin{enumerate}
\item Regression: It represents training the regression network only on RGB images and without any depth regularizer. 
\item Regression + DR + DGGAN: We learne the depth-regularized regression network using RGB images with the depth maps generated by DGGAN.
\item Regression + DR + true depth map: We derive the depth-regularized regression network using RGB images with their paired true depth maps.
\end{enumerate}

To measure the effectiveness of the generated depth maps, we compare settings Regression and Regression + DR + DGGAN. 
As illustrated in Table~\ref{table:3Dresult}, using the generated depth map significantly boosts the performance of the model in Regression. 
The AUC $20$\_$50$ is improved by \textbf{0.043}, \textbf{0.024}, \textbf{0.011} on the RHD, STB, and MHP datasets, respectively. 
The EPE mean is also considerable reduced by \textbf{13.2\%} and \textbf{19.7\%} and \textbf{7.3\%} on the RHD, STB and MHP datasets respectively.


To compare the generated depth map with the real depth maps, we conduct two more experiments.
Comparing results of Regression + DR + true depth map and Regression + DR + DGGAN shows that the generated depth maps are a key factor of performance boosting. 
On the RHD dataset, training with the generated depth maps is only slightly worse than the true RHD depth maps by $0.02$ in AUC $20$\_$50$ and $1$ mm in EPE mean. 
However, on the STB dataset, the results of training with generated depth maps are even better than training with the real depth maps (by $0.006$ in AUC $20$\_$50$ and $0.94$ mms in EPE mean).
This result is probable due to the fact that the depth maps collected from depth sensors are less stable and noisier than the depth maps collected from a $3$D simulator. 
By training the DGGAN with unpaired high-quality depth maps from RHD, our generator can potentially reduce the noise, and further benefit the training in the hand pose estimation module. 
It is worth noting that Regression + DR + true depth map requires the paired depth and RGB image.

In addition to the quantitative analysis, Figure~\ref{fig:RHDdepthcmp} and Figure~\ref{fig:STBdepthcmp} provide some examples for visual comparison between the generated and true depth maps on the RHD and STB datasets, respectively. 
We can see that the generated depth maps are visually very similar to the ground-truth ones.
\subsection{Comparison with State-of-the-arts}
We select the state-of-the-art approaches~\cite{boukhayma20193d,cai2018weakly, panteleris2018using,Spurr_2018_CVPR,zhang20163d,mueller2018ganerated,zb2017hand} for comparison. 
%
The comparison results are reported in Figure~\ref{fig:teaser} and Table~\ref{table:stoa}. 
As shown in Figure~\ref{fig:teaser} and Table~\ref{table:stoa}, our approach outperforms all existing state-of-the-art methods. 
Although the results of the method by Cai~\etal~\cite{cai2018weakly} come close to ours, we emphasize that our DGGAN has an crucial advantage of \emph{not} requiring any paired RGB and depth images.

\section{Conclusion}
\label{sec:con}
The lack of large-scale datasets of paired RGB and depth images is one of the major bottlenecks for improving $3$D hand pose estimation. 
To address this limitation, we propose a conditional GAN-based  model called DGGAN to bridge the gap between RGB images and depth maps. 
DGGAN synthesizes depth maps from RGB images to regularize the $3$D hand pose prediction model during training, eliminating the need of paired RGB images and depth maps conventionally used to train such models. 

The proposed DGGAN is integrated into a $3$D hand pose prediction framework, and is trained end-to-end together for $3$D pose estimation.
DGGAN not only generates more realistic hand depth images, which can be used in many other applications such as $3$D shape estimation but also results in significant improvement in $3$D hand pose estimation, achieving new state-of-the-art results. 

\paragraph{Acknowledgement.}
This work
was supported in part by Ministry of Science and Technology
(MOST) under grants MOST 107-2628-E-001-005-MY3 and MOST 108-2634-F-007-009.

{\small
\bibliographystyle{ieee}
\bibliography{egbib}
}

\end{document}